\documentclass[twocolumn,english]{article}
\usepackage[T1]{fontenc}
\usepackage[latin9]{inputenc}
\usepackage{float}
\usepackage{units}
\usepackage{graphicx}

\floatstyle{ruled}
\newfloat{algorithm}{tbp}{loa}
\floatname{algorithm}{Algorithm}

\newtheorem{definition}{Definition}
\newtheorem{theorem}{Theorem}

\newtheorem{lemma}{Lemma}

\newcommand{\qed}{{\setlength{\fboxsep}{0pt}
\framebox[7pt]{\rule{0pt}{7pt}}}}

\newenvironment{proof}{{\bf Proof:}\/}{\vskip -0.5cm \hspace*{\fill} \qed \vskip 0.0in}

\usepackage{icml2008}
\usepackage{mlapa}
\usepackage{url}

\usepackage{babel}

\begin{document}
\global\long\def\spn{\mbox{span}}

\global\long\def\con{{\bf contract}}

\global\long\def\lab{l}

\global\long\def\cla{c}

\global\long\def\rat{\alpha}

\global\long\def\ins{n}

\global\long\def\alg{\mathbf{alg}}

\twocolumn[
\icmltitle{Efficient Human Computation: the Distributed Labeling Problem}
\icmlauthor{}{}
\icmladdress{{\bf Keywords}: supervised learning, theory}
\icmlauthor{Ran Gilad-Bachrach}{ran.gilad-bachrach@intel.com}
\icmladdress{Intel Research Israel Lab}
\icmlauthor{Aharon Bar-Hillel}{aharon.bar-hillel@intel.com}
\icmladdress{Intel Research Israel Lab}
\icmlauthor{Liat Ein-Dor}{liat.ein-dor@intel.com}
\icmladdress{Intel Research Israel Lab}

\vskip 0.3in
]
\begin{abstract}
Collecting large labeled data sets is a laborious and expensive task,
whose scaling up requires division of the labeling workload between
many teachers. When the number of classes is large, miscorrespondences
between the labels given by the different teachers are likely to occur,
which, in the extreme case, may reach total inconsistency. In this
study we describe how globally consistent labels can be obtained,
despite the absence of teacher coordination, and discuss the possible
efficiency of this process in terms of human labor. We define a notion
of label efficiency, measuring the ratio between the number of globally
consistent labels obtained and the number of labels provided by distributed
teachers. We show that the efficiency depends critically on the ratio
$\alpha$ between the number of data instances seen by a single teacher,
and the number of classes. We suggest several algorithms for the distributed
labeling problem, and analyze their efficiency as a function of $\alpha$.
In addition, we provide an upper bound on label efficiency for the
case of completely uncoordinated teachers, and show that efficiency
approaches $0$ as the ratio between the number of labels each teacher
provides and the number of classes drops (i.e. $\alpha\to0$).
\end{abstract}

\section{Introduction}

As applications of machine learning mature, larger training sets are
required both in terms of the number of training instances and the
number of classes considered. In recent years we have witnessed this
trend for example in vision related tasks such as object class recognition
or detection \cite{ghp07,evwwz07,rtmf05}. Specifically for object
class recognition, current data sets such as the Caltech-256 \cite{ghp07}
include tens of thousands of images from hundreds of classes. Collecting
consistent data sets of this size is an intensive and expensive task.
Scaling up naturally leads to a distributed labeling scenario, in
which labels are provided by a large number of weakly coordinated
teachers. For example, in the Label-me system \cite{rtmf05} the labels
are contributed by dozens of researchers, while in the ESP game \cite{vonAhn06}
labels are supplied by thousands of uncoordinated players.

As we turn toward distributed labeling, several practical considerations
emerge which may disrupt the data integrity. In general, while it
is reasonable to believe that a single teacher is relatively self-consistent
(though not completely error-free), this is not the case with multiple
uncoordinated teachers. Different teachers may have differences in
their labeling systems due to several causes. First, different teachers
may use different words to describe the same item class. For example,
one teacher may use the word ``truck'' while the other uses ``lorry''
to describe the same class. Conversely, the same word may be used
by two teachers to describe two totally different classes, hence one
teacher may use ``greyhound'' to describe the breed of dog while the
other uses it to describe the C-2 navy aircraft. Similar problems
occur when different teachers label the data with different abstraction
levels, so one generalizes over all dogs, while the other discriminates
between a poodle, a Labrador and etc. Finally, teachers often do not
agree on the exact demarcation of concepts, so a chair carved in stone
may be labeled as a ``chair'' by one teacher, while the other describes
it as ``a rock''. All these phenomena become increasingly pronounced
as the number of classes is increased, thus their neglect essentially
leads to a severe decrease in label purity and consequently in learning
performance.

In this paper we study the cost of obtaining globally consistent labels,
while focusing on a specific distributed labeling scenario, in which
only some of the difficulties described above are present. To enforce
the distributed nature of the problem, we assume that a large data
set with $\ins$ examples is to be labeled by a set of uncoordinated
teachers, where each teacher agrees to label at most $\lab\ll\ins$
data points. While there is a one-to-one correspondence between the
classes used by the different teachers, we assume that their labeling
systems are entirely uncoordinated, so a class labeled as ``duck''
by one teacher may be labeled as a ``goat'' by another. In later stages
of this paper, we relax this assumption, and consider a case in which
partial consistency exists between the different teachers. Both scenarios
are realistic in various problem domains. Consider for example a security
system for which we have to label a large set of face images, including
thousands of different people. Since teachers are not familiar with
the persons to be labeled, the names they give to classes are entirely
un-coordinated. The case of a partial consistency is exemplified in
distributed labeling of flower images: the layman can easily distinguish
between many different kinds of flowers but can name only a few.

The difficulties of ``one-to-many'' label correspondence between teachers
and concept demarcation disagreements are not met by our current analysis,
which focuses on the preliminary difficulties of distributed labeling.
Another related scenario, to which our analysis can be extended relatively
easily, is the case in which the initial data is labeled by uncoordinated
teachers right from the start. Consider for example, the task of unifying
images labeled in a site like Flickr%
\footnote{http://www.flickr.com/%
} into a meaningful large training data set. Our suggested algorithms
and analysis apply to this case with minor modifications.

\subsection{Relevant literature}

In the active learning framework \cite{cal90} and the experimental
design framework (see e.g., \cite{ad92}), the goal is to minimize
the number of queries for labels (or experiments conducted) while
learning a target concept. It has been shown \cite{fsst95} that a
careful selection of queries can lead to an exponential reduction
in the number of labels needed. This line of research is motivated
by the costly and cumbersome process of obtaining labels for instances.
We share this motivation but argue that the problem is not merely
the quantity of labels but also the quality and the consistency of
the labels that should be treated in the data collection process.

The problem of quality of labels, i.e., learning with noise, has been
addressed extensively in the machine learning literature (see e.g.,
\cite{Decatur95}). In this line of work it is assumed that the teacher
does not always provide the true instance labels. The severity of
noise ranges from adversarial noise, in which the teacher tries to
prevent the learning process by providing inaccurate labels, to the
more benign random classification noise. While the inconsistency between
uncoordinated teachers can be regarded as some form of label noise,
it has unique characteristics and its treatment is hence different
from the other sources of noise mentioned. Specifically, as long as
each teacher is noise-free and self-consistent, we are able to eliminate
the noise completely and achieve certain labels.

The scenario of distributed labeling with uncoordinated teachers was
considered in the ``equivalence constraints'' framework \cite{bhsw05}.
When learning with equivalence constraints, the learner is presented
with pairs of instances and the annotation suggests whether they share
the same class or not. The authors conjectured that as the number
of classes increase, the labeling effort required to coordinate the
labels from different teachers becomes prohibitive. We prove this
conjecture in Theorem~\ref{th: upper bound on f^*}. Alternatively,
equivalence constraints can be used as a direct supervision for the
learning algorithm. Indeed, \cite{bw03} proved that a concept class
is learnable with equivalence constraints if it is learnable from
labels, so this alternative has some appeal. 

\comment{In this work we provide mathematical support to the claim
regarding the high cost of achieving globally consistent labels as
the number of classes rises. However, we do not tend to conclude that
using equivalence constraints for learning is preferable in this case.
The equivalence constraints gathered in this scenario are mostly negative
constraints (stating that two points are not from the same class),
which have low information content and are often computationally hard
to use. While this issue should be addressed more carefully, it seems
that gathering enough useful constraints is likely to require labeling
effort comparable to the effort needed for consistent label production.
Finally, algorithms for supervised learning from equivalence constraints
are not mature yet.}

\subsection{The distributed labeling problem\label{sub:The-distributed-labeling}}

In the distributed labeling task we have to reveal the labels of $\ins$
instances $\left\{ x_{1},\ldots,x_{\ins}\right\} $. We assume that
there exist ``true'' labels $y_{1},\ldots,y_{n}$ (with $y_{j}=y(x_{j})$)
and the distributed labeling algorithm should return $\bar{y}_{1},\ldots,\bar{y}_{\ins}$
such that $\bar{y}_{i}=\bar{y}_{j}$ if and only if $y_{i}=y_{j}$.
We denote the number of classes by $\cla$, and assume that each teacher
is willing to label only $\lab=\cla\rat$ instances where $\lab,\cla\ll\ins$.
Throughout this paper we assume that the labels provided by teachers
are consistent with the true labels in the sense that for any teacher
$t$ and any pair of instances $x_{i},x_{j}$ \begin{equation}
\left[t\left(x_{i}\right)=t\left(x_{j}\right)\right]\Longleftrightarrow\left[y_{i}=y_{j}\right]\,\,\,.\label{eq:class consistent}\end{equation}

where $t\left(x_{}\right)$ is the label given by teacher $t$ to
instance $x$. However, apart from section~\ref{sec:Learning-with-name-consistent},
we assume no inter-teacher consistency with respect to class names,
i.e., teachers may disagree on the names of the different classes.
To measure the competence of different algorithms for combining the
labels of the different teachers we define the following:

\begin{definition} Denote by $\sigma=\left\{ x_{i},y_{i}\right\} _{i=1}^{\ins}$
an input sequence of $\ins$ points with the labels $y_{i}\in\left\{ 1,..,\cla\right\} $.
A distributed labeling algorithm $\alg$ is $f\left(\rat,\alg\right)$
\emph{efficient} if\[
f\left(\rat,\alg\right)=\lim_{\cla\rightarrow\infty}\lim_{\ins\rightarrow\infty}\frac{\ins}{\sup_{\sigma}\left(\mathbf{labels}\left(\alg,\sigma,\cla\alpha\right)\right)}\]

where $\mathbf{labels}\left(\alg,\sigma,\lab\right)$ is the average
(over the internal randomness of the algorithm) number of human-generated
labels the algorithm $\alg$ uses to label the sequence $\sigma$,
where each teacher is willing to label $l$ examples.

\end{definition}

Clearly, if no structural assumptions are made on true labels then
$f\left(\alpha,\alg\right)$ is bounded by $1$ from above. We denote
by $f^{*}\left(\alpha\right)$ the optimal efficiency for a given
$\alpha$. I.e., $f^{*}\left(\alpha\right)=\sup_{\alg}f\left(\alpha,\alg\right)$.

\subsection{Main results}

In section \ref{sec:Label-Efficiency} we present several algorithms
for solving the distributed labeling problem. The first algorithm
presented is the \emph{contract the connected components} $(C^{3})$
algorithm. We show that this simple algorithms has efficiency of $1-\nicefrac{\left(1-\exp\left(-\alpha\right)\right)}{\alpha}$.
We then improve this algorithm with the \emph{representatives algorithm}
and prove its efficiency to be better than the efficiency of the previous
algorithm. In section~\ref{sec:The-optimal-efficiency} we present
an upper bound on the achievable efficiency. We show that $f^{*}\left(\alpha\right)\leq\min\left(\nicefrac{2\alpha}{\left(1+\alpha\right)},1\right)$.
In section~\ref{sec:Learning-with-name-consistent} we study a relaxed
version of the distributed labeling problem in which there exists
some consistency between the different teachers. Thus, with some probability
$p$ two teachers will agree on the name of a given class. In this
setting, we present a revised version of the $C^{3}$ algorithm and
show its efficiency to be $1-\frac{1-\exp\left(-\alpha\right)}{\alpha-\exp\left(-\alpha\right)+\exp\left(-\alpha\left(1-p\right)\right)}$.

\section{Label-efficient algorithms\label{sec:Label-Efficiency}}

\begin{figure}
\includegraphics[width=0.95\columnwidth]{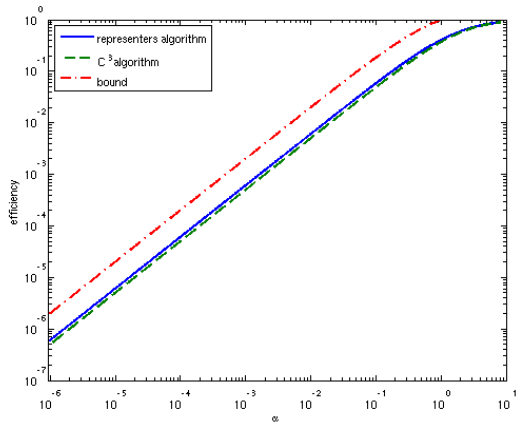}

\caption{\label{fig:The-efficiency}The efficiency (Y-axis) of the $C^{3}$
algorithm and the anchor algorithm are plotted together with the bound
on the optimal efficiency (Theorem~\ref{th: upper bound on f^*})
for different values of $\alpha$ (X-axis).}

\end{figure}

As described in \ref{sub:The-distributed-labeling}, we assume in
this section that the name each teacher assigns to a class is meaningless.
Therefore, the best we can hope for is to break the $n$ instances
into $c$ classes such that any pair of points share the same class
label if and only if all teachers give these two points the same label.
In this section we suggest two algorithms for this task. The bounds
obtained for these algorithms are presented in Figure~\ref{fig:The-efficiency}.

\subsection{The \textit{Contract} the \textit{Connected Components} ($C^{3}$)
algorithm}

The first algorithm we consider is the \textit{Contract} the \textit{Connected
Components} ($C^{3}$) algorithm presented in Algorithm~\ref{alg:c3}.
The idea behind this algorithm is to build a graph whose nodes are
sets of equivalent instances. Whenever we find that two nodes share
the same label, we contract them into a single node. On the other
hand, whenever we find that two nodes do not share the same label,
we generate an edge between them. The algorithm ends when the remaining
graph is a clique. At this point, each of the nodes is assigned with
a unique label. These labels propagate to all the points to be labeled,
since each point is associated with a single node in the clique.

\begin{algorithm}[t]

\caption{\textbf{\label{alg:c3}}The\textbf{ }\textit{Contract} the \textit{Connected
Components} ($C^{3}$)\textbf{ }algorithm}

\textbf{input}: $\ins$ unlabeled instances $x_{1},\ldots,x_{\ins}$

\textbf{output}: a partition of $x_{1},\ldots,x_{\ins}$ into classes
according to the true labels
\begin{enumerate}
\item Let $G$ be the edge-free graph whose vertexes are $x_{1},\ldots,x_{\ins}$.
\item While $G$ is not a clique

\begin{enumerate}
\item pick $l$ random nodes $U=\left\{ x_{i_{1}},\ldots,x_{i_{l}}\right\} $
which are not a clique from $G$.
\item send $U$ to a teacher and receive $y_{i_{1}},\ldots,y_{i_{l}}$.
\item for every $1\leq r<s\leq\lab$ do

\begin{enumerate}
\item if $y_{i_{r}}=y_{i_{s}}$ then contract the vertices $x_{i_{r}}$
and $x_{i_{s}}$ in the graph $G$.
\item if $y_{i_{r}}\neq y_{i_{s}}$ then add the edge $\left(x_{i_{r}},x_{i_{s}}\right)$
to the graph $G$.
\end{enumerate}
\end{enumerate}
\item Mark each vertex in $G$ with a unique number from $\left[1\ldots\cla\right]$.
\item For every vertex in $G$, propagate its label to all the nodes that
were contracted into this vertex.
\end{enumerate}

\end{algorithm}

The correctness of the algorithm is straightforward due to the self-consistency
of the teachers. In Theorem~\ref{th: c3} we show the label efficiency
of the $C^{3}$ algorithm to be $1-\left(1-\exp\left(-\alpha\right)\right)/\alpha$
where $\alpha=l/c$. The main idea behind the analysis is to study
the expected number of contractions in each iteration.

\begin{theorem}\label{th: c3} The label efficiency of the $C^{3}$
algorithm is lower-bounded by \[
1-\frac{1}{\alpha}\left(1-\exp\left(-\alpha\right)\right)\,\,\,.\]

\end{theorem}

Before proving the theorem, we present a lemma in which the contraction
rate associated with a single teacher is bounded.

\begin{lemma}\label{le: Q function} Assume a teacher labels $\lab$
random example ($\lab\rightarrow\infty)$ from $\cla=\nicefrac{\lab}{\alpha}$
different classes. The expected number of unique labels that the teacher
will give to the $l$ instances is at most $\lab$ times $Q\left(\alpha\right)$
where \[
Q\left(\alpha\right)=\frac{1}{\alpha}\left(1-\exp\left(-\alpha\right)\right)\,\,\,.\]

\end{lemma}

Note that the number of unique labels is exactly the number of nodes
that will be left after contracting the $\lab$ instances.

\begin{proof} Assume that the probability for seeing each of the
classes is $p_{i}$. The result follows from the following:\begin{eqnarray}
\lefteqn{E\left[\mbox{number of unique labels}\right]}\nonumber \\
 & = & \cla-E\left[\mbox{number of labels not seen}\right]\nonumber \\
 & = & c-\sum_{i}\left(1-p_{i}\right)^{l}\label{eq:pi}\\
 & \leq & \cla-\cla\left(1-\frac{1}{\cla}\right)^{\lab}\nonumber \\
 & = & \cla\left(1-\exp\left(-\alpha\right)\right)\label{eq:exponent}\\
 & = & l\cdot\frac{1}{\alpha}(1-\exp\left(-\alpha\right))\,\,\,.\nonumber \end{eqnarray}
The correctness of (\ref{eq:exponent}) follows since we are assuming
that $\lab,\cla\rightarrow\infty$ while $\alpha=\lab/\cla$ is constant.

\end{proof}

\begin{proof} \emph{(of Theorem~\ref{th: c3}) }At each round of
the $C^{3}$ algorithm, $\lab$ elements are sent to be labeled by
a teacher. From Lemma~\ref{le: Q function} we have that the number
of remaining elements is on avarage at most $\lab Q\left(\alpha\right).$

Therefore, the expected number of rounds the algorithm will make until
finished is \[
\frac{\ins}{\lab\left(1-\frac{1}{\alpha}(1-\exp\left(-\alpha\right))\right)}\,\,\,.\]
Note that the number in the denominator is the expected number of
removed elements at each round. Thus, the number of labels used is
\[
\frac{n}{\left(1-\frac{1}{\alpha}(1-\exp\left(-\alpha\right))\right)}\,\,\,.\]
Plugging this number into the definition of label efficiency gives
the desired result.\end{proof}

\subsection{The representatives algorithm \label{sub:The-representative-algorithm}}

Each teacher provides us with two types of information sources. One
is positive equivalence constraints, i.e., the knowledge that two
instances share the same label. The other is negative equivalence
constraints, i.e., the knowledge that two instances do not share the
same label. While the $C^{3}$ algorithm is very effective in using
positive equivalence constraints, it makes very little use of negative
equivalence constraints. The \emph{representatives algorithm} (Algorithm~\ref{alg:The-representative-Algorithm})
tries to exploit this type of information as well. The main idea behind
this algorithm is first to find all the points that belong to certain
classes. Once we know that the remaining points do not belong to any
of these classes, we are left with a problem with fewer instances
and fewer potential classes and thus an ``easier one''.

In order to detect all the points belonging to a certain class we
use \emph{representatives}. A \textit{representatives set} is a set
of $c$ instances $\left\{ x_{i_{1},..,}x_{i_{\cla}}\right\} $such
that for each class there is exactly one member (representative) of
the class in the representatives set. Finding a representatives set
is a simple task and can be done without affecting the overall efficiency,
since its label complexity does not depend on $n$. Therefore, for
the sake of simplicity we assume that the representatives set is given
in advance. We further assume that we know the probability of each
representative class. This information too can be easily estimated
from data without jeopardizing efficiency.

\begin{algorithm}[t]
\caption{\label{alg:The-representative-Algorithm}The Representatives Algorithm}

Inputs: 
\begin{itemize}
\item $\ins$ unlabeled instances, $x_{1}^{},\ldots,x_{n}^{}$ 
\item a set $a_{1},\ldots,a_{\cla}$ of representatives such that $a_{i}\in\left\{ x_{1},\ldots,x_{n}^{}\right\} $
\item a list of probabilities $p_{1},\ldots,p_{\cla}$ such that $p_{i}$
is the probability of seeing an instance from the class of $a_{i}$. 
\end{itemize}
Outputs: a partition of the $n$ points into $c$ label classes
\begin{enumerate}
\item Reorder the representatives and the $p_{i}$'s such that $p_{1}\geq p_{2}\geq\ldots\geq p_{\cla}$.
\item Let{*} $\beta\in\left(0,1\right)$
\item Partition the set of representatives into $r$ sets $S_{0},\ldots,S_{r-1}$
classes such that $S_{i}=\left\{ a_{i\beta l+1},\ldots,a_{(i+1)\beta l}\right\} $.
\item Let $G$ be the edge free graph whose vertices are $x_{1},\ldots,x_{\ins}$.
\item While $G$ is not empty

\begin{enumerate}
\item For $i=0\ldots r-1$\label{enu:anchor-for}

\begin{enumerate}
\item Partition the remaining points in the graph into sets of size $\left(1-\beta\right)l$.
\item For each subset of $\left(1-\beta\right)l$ points:

\begin{enumerate}
\item send these points together with $S_{i}$ to a teacher.
\item contract the graph according to the labels returned by the teacher.
\end{enumerate}
\item For every $a_{j}\in S_{i}$

\begin{enumerate}
\item label $a_{j}$ with the label $j$, and propogate this label.
\item remove $a_{j}$ from $G$.
\end{enumerate}
\end{enumerate}
\end{enumerate}
\end{enumerate}
~

{*} Choose $\beta$ to optimize the bound in Theorem~\ref{th: anchor}.
\end{algorithm}

$\beta$ is the proportion of representatives in the $l$ instances
each teacher labels. Note that when $\beta=0$, the representative
algorithm is essentially the same as the $C^{3}$ algorithem. However,
when $\beta>0$, we use the fact that after all the points were compared
against a certain representative, we are guaranteed to have found
all the points with the same label as this representative, and thus
we can eliminate this class.

\begin{theorem} \label{th: anchor}The label efficiency of the representative
algorithm is lower-bounded by\[
\frac{\left(1-\beta\right)\left(1-q\right)^{2}}{1-q-\frac{q}{r}\left(1-q^{r}\right)}\]
 where $r=\frac{\cla}{\beta l}=\frac{1}{\alpha\beta}$ is the number
of sets in the partition of the representatives into $\beta l$ sets
and%
\footnote{The $Q$ function is defined in Lemma~\ref{le: Q function}.%
} $q=Q\left(\alpha\left(1-\beta\right)\right)=\frac{1-\exp\left(-\alpha\left(1-\beta\right)\right)}{\alpha\left(1-\beta\right)}$
.

\end{theorem}

\begin{proof} In each round of step \ref{enu:anchor-for} we break
$G$ into $\left|G\right|/\left(l\left(1-\beta\right)\right)$ parts
and thus use $\left|G\right|/\left(1-\beta\right)$ labels. Therefore,
we need only to estimate the size of $G$ after each round. Denote
the number of vertices in $G$ at the beginning of the round $i$
by $g_{i}$. In order to bound $g_{i}$ we should consider how it
is affected by two ingredients: first the contraction which happen
in the same fashion as it happens in the $C^{3}$ algorithm and the
complete elimination of classes $1,..,i\beta\lab$.

We use Lemma~\ref{le: Q function} to analyze the contraction rate.
Each teacher sees $l\left(1-\beta\right)$ instances which are not
representers of some classes. These instances come from $\cla-i\beta\lab$
different classes and thus, from Lemma~\ref{le: Q function} the
contraction rate is \[
Q\left(\frac{\lab\left(1-\beta\right)}{\cla-i\beta\lab}\right)=Q\left(\frac{\alpha\left(1-\beta\right)}{1-i\alpha\beta}\right)\,\,\,.\]
Out of the remaining points, all the points which are being represented
in $S_{i}$ are eliminated. Due to the reordering of the $p_{i}$s,
these points are at least a fraction of $\nicefrac{1}{\left(r-i\right)}$
of the remaining points. Thus\begin{eqnarray*}
g_{i+1} & \leq & g_{i}\frac{r-\left(i+1\right)}{r-i}Q\left(\frac{\alpha\left(1-\beta\right)}{1-i\alpha\beta}\right)\\
 & = & \ins\left(\prod_{j=0}^{i}\frac{r-\left(j+1\right)}{r-j}\right)\left(\prod_{j=0}^{i}Q\left(\frac{\alpha\left(1-\beta\right)}{1-j\alpha\beta}\right)\right)\\
 & = & n\left(1-\frac{i+1}{r}\right)\prod_{j=0}^{i}Q\left(\frac{\alpha\left(1-\beta\right)}{1-j\alpha\beta}\right)\,\,\,.\end{eqnarray*}

The number of labels used in all the rounds is therefore\begin{eqnarray}
\lefteqn{\sum_{i=0}^{r-1}\frac{g_{i}}{\left(1-\beta\right)}\le}\nonumber \\
 &  & \frac{n}{1-\beta}\sum_{i=0}^{r-1}\left(1-\frac{i}{r}\right)\prod_{k=0}^{i-1}Q\left(\frac{\alpha(1-\beta)}{1-k\alpha\beta}\right)\label{eq:representers not bounded}\\
 & \leq & \frac{n}{1-\beta}\sum_{i=0}^{r-1}\left(1-\frac{i}{r}\right)Q\left(\alpha\left(1-\beta\right)\right)^{i}\nonumber \\
 & = & \frac{n\left(1-q-\frac{q}{r}\left(1-q^{r}\right)\right)}{\left(1-\beta\right)\left(1-q\right)^{2}}\nonumber \end{eqnarray}

where (\ref{eq: Q monotone}) is due to the monotonicity of the $Q$
function. Using the last expression in the efficiency definition completes
the proof.\end{proof}

The expression obtained in theorem \ref{th: anchor} can be computed
numerically for any value of $\alpha,\beta$ and so it can be used
to optimize $\beta$ for a given $\alpha$. When the optimal $\beta$
is used, the representers algorithm outperforms the $C^{3}$ algorithm
as seen in Figure~\ref{fig:The-efficiency}.

\comment{Furthermore, this expression generalizes the efficiency
expression of theorem \ref{th: c3} obtained for the $C^{3}$ algorithm,
as can be seen by taking $\beta=0$. Furthermore, if $\alpha\rightarrow0$
and $\beta=\nicefrac{1}{3}$ the bound for the efficiency obtained
is $\nicefrac{2}{3}\alpha$ which is better than the efficiency of
the $C^{3}$ algorithm in this case which is $\nicefrac{1}{2}$.}

\section{The optimal efficiency\label{sec:The-optimal-efficiency}}

In the previous section we studied the efficiency of several algorithms.
In the current section we study the efficiency of the optimal algorithm.
That is, we study the function\[
f^{*}\left(\alpha\right)=\sup_{\alg}f\left(\alpha,\alg\right)\,\,\,.\]
We give an upper bound on $f^{*}\left(\alpha\right)$ showing that
algorithms cannot have an efficiency greater than $\min\left(1,\nicefrac{2\alpha}{\left(1+\alpha\right)}\right)$.
This bound asserts that the labeling problem is not trivial in the
sense that it is not always possible to achieve efficiency 1. Moreover,
the problem becomes hard in the limit of $\alpha\to0$, as the efficiency
drop linearly with $\alpha$ in this region. Comparing the bound shown
here and the efficiency of the algorithms presented in previous sections,
one can see that there is still a significant gap between the achieved
and the (maybe) achievable.

\begin{theorem}\label{th: upper bound on f^*} Let $f^{*}\left(\alpha\right)$
be the best achievable efficiency for a given $\alpha$ then\[
f^{*}\left(\alpha\right)\leq\min\left(1,\frac{2\alpha}{1+\alpha}\right)\,\,\,.\]

\end{theorem}

\begin{proof} Fix $\ins$ and $\cla$ and assume $\lab=\alpha\cla$.
If $\alpha>1$ then the required bound is trivial since efficiency
cannot exceed $1$. Therefore, we are only interested in the cases
where $\alpha<1$. Let $\alg$ be a distributed labeling algorithm.
For each of the $n$ instances we choose a class label uniformly and
independently from the $\cla$ possible labels. We analyze the expected
number of teacher calls needed before the class assignments are found. 

Fix an instance $x$, we first analyze the expected number of teacher
calls (in which $x$ participates) before it is first contracted with
some other point. Assume that $x$ has $i$ edges in the graph $G$,
i.e., there are $i$ instances for which it is known that $x$ does
not share its label. If $x^{\prime}$ is a different point than $x$,
the probability that they share the same label is at most $\nicefrac{1}{\left(\cla-i\right)}$.
To see this, note that for any legal label assignment to $G\setminus\left\{ x\right\} $,
there are at least $\cla-i$ uplifts of this assignment to $G$. 

Let $P\left(i\right)$ be the probability that $x$ is contracted
at least once during its first $i$ comparisons to other instances.
We claim that $P\left(i\right)\leq\nicefrac{i}{c}$ for all $1\le i\le\cla$.
Clearly, $P(0)=0$. The proof is by induction. For $i=1$, clearly
the probability for contraction with the first point $x$ is compared
against is $\nicefrac{1}{c}$. Note that \begin{eqnarray*}
\lefteqn{P\left(i+1\right)}\\
 & = & P\left(i\right)+\left(1-P\left(i\right)\right)\Pr\left[\con\,\mbox{at step }i+1\right]\\
 & \leq & P\left(i\right)+\left(1-P\left(i\right)\right)\frac{1}{\cla-i}\\
 & \leq & \frac{i}{\cla}\left(1-\frac{1}{\cla-i}\right)+\frac{1}{\cla-i}=\frac{i+1}{\cla}\,\,\,.\end{eqnarray*}

In the previous calculation, we assumed that $x$ is compared to other
points one at a time. However, the teachers label $\lab$ instances
at a time, thus whenever $x$ is sent to a teacher, it is compared
against $\lab-1$ points. Note that an instance keeps being sent to
teachers at least until it is first unified. Therefore, the number
of teachers that will have to label $x$ until its label is discovered,
is at least the total number of teachers that will have to label x
until it is unified at least once with another instance. From this
we obtain the following lower bound for the expected number of teachers
that see x: \begin{eqnarray*}
\lefteqn{E\left[\mbox{number of teachers that see }x\right]}\\
 & = & \sum_{j}\Pr\left[\mbox{number of teachers}\geq j\right]\\
 & = & \sum_{j}\left(1-\Pr\left[\mbox{number of teachers}<j\right]\right)\\
 & \ge & \sum_{j=1}^{\nicefrac{\left(\cla-1\right)}{\left(\lab-1\right)}}\left(1-P\left(\left(j-1\right)\left(\lab-1\right)\right)\right)\\
 & \geq & \sum_{j=1}^{\nicefrac{\left(\cla-1\right)}{\left(\lab-1\right)}}\left(1-\frac{\left(j-1\right)\left(\lab-1\right)}{\cla}\right)\\
 & = & \frac{\cla-1}{\lab-1}-\frac{1}{2}\left(\frac{\cla-l}{\lab-1}\right)\left(\frac{\cla-1}{\cla}\right)\,\,\,.\end{eqnarray*}

The efficiency can be derived from this term \begin{eqnarray*}
\lefteqn{f^{*}\left(\alpha\right)}\\
 & \le & 1/\lim_{\cla\rightarrow\infty}\left(\frac{\cla-1}{\lab-1}-\frac{1}{2}\left(\frac{\cla-l}{\lab-1}\right)\left(\frac{\cla-1}{\cla}\right)\right)\\
 & = & 1/\left(\frac{1}{\alpha}-\frac{1}{2}\left(\frac{1}{\alpha}-1\right)\right)=\frac{2\alpha}{1+\alpha}\,\,\,.\end{eqnarray*}

\end{proof}

\section{Learning with name-consistent teachers\label{sec:Learning-with-name-consistent}}

In previous sections we assumed that class names used by different
teachers are totally uncoordinated, so naming conventions of one teacher
are meaningless to the other. While this scenario may occur (like
in the 'face labeling' task mentioned in the introduction), in most
cases this assumption is too pessimistic. It is more reasonable to
assume that some level of agreement regarding class names exist, though
this agreement is partial and not perfect. In this section we assume
that there exist $0\leq p\leq1$ such that with probability $p$ over
the choice of a random teacher $t$ and class $j$, the teacher uses
the true global class name $j$ as the class label:

\begin{eqnarray}
\forall x\,\,\,\,\,\,\Pr_{t}\left(t\left(x\right)=y\left(x\right)\right) & \geq & p\,\,\,.\label{eq:name consistent}\end{eqnarray}
 We assume some sort of a probability measure over the teachers and
the classes. If the pool of teachers is finite, it can be the uniform
distribution, and otherwise we assume that whenever we need another
teacher to label some instances, the teacher will be such that (\ref{eq:name consistent})
is true. Notice that we also keep our previous assumption that all
the teachers are class consistent in the sense of (\ref{eq:class consistent}).

When $p=1$ the assumption (\ref{eq:name consistent}) means that
all the teachers use the same global naming system , i.e. $t(x_{j})=y_{j}$
for all $t,j$. In this case the labeling problem is trivial, and
it is easy to obtain label efficiency of $1$ simply by splitting
the instances between different teachers. On the other hand, when
$p$ is very small, there is no name consistency and the situation
boils down to the scenario studied in Section~\ref{sec:Label-Efficiency}.
Therefore, we will now focus on studying name consistency in the general
case when $p\in\left(0,1\right)$.

The algorithm we present to address this situation is the \textit{Consistently
Contract} the \textit{Connected Components} ($C^{4}$) (Algorithm~\ref{alg:The-Consistently-Connect}).
The difference between the $C^{4}$ algorithm and the $C^{3}$ algorithm
is that the $C^{4}$ algorithm sends teachers instances that were
previously given the same label by some other teachers.

\begin{algorithm}[t]
\caption{\textbf{\label{alg:The-Consistently-Connect}The Consistently Contract
the Connected Components ($C^{4}$) algorithm}}

Input: $\ins$ unlabeled instances $x_{1},\ldots,x_{\ins}$

Output: a partition of $x_{1},\ldots,x_{\ins}$ into classes according
to the true labels
\begin{enumerate}
\item Let $G$ be the edge free graph whose vertices are $x_{1},\ldots,x_{\ins}$.
\item Label each vertex with $0$.
\item While $G$ is not a clique

\begin{enumerate}
\item pick $l$ random nodes $U=\left\{ x_{i_{1}},\ldots,x_{i_{l}}\right\} $
from $G$ such that all these nodes have the same label.
\item send $U$ to a teacher and receive $y_{i_{1}},\ldots,y_{i_{l}}$.
\item for every $1\leq r\leq l$ , label $x_{i_{r}}$ with the label $y_{i_{r}}$.
\item for every $1\leq r<s\leq\lab$ do

\begin{enumerate}
\item if $y_{i_{r}}=y_{i_{s}}$ then contract the vertices $x_{i_{r}}$
and $x_{i_{s}}$ in the graph $G$.
\item if $y_{i_{r}}\neq y_{i_{s}}$ then add the edge $\left(x_{i_{r}},x_{i_{s}}\right)$
to the graph $G$.
\end{enumerate}
\end{enumerate}
\item Mark each vertex in $G$ with a unique number.
\item For every vertex in $G$ propagate its label to all the nodes that
were contracted into this vertex.
\end{enumerate}

\end{algorithm}

The $C^{4}$ algorithm differs from the $C^{3}$ algorithm in using
the labels for selecting better candidates for sending to the same
teacher. However, note that we still declare the equivalence of two
instances only when a single teacher labels both with the same label.
Therefore, due to the class consistency (\ref{eq:class consistent})
the correctness of the algorithm is guaranteed. We now turn to proving
its efficiency.

\begin{theorem} The label efficiency of the $C^{4}$ algorithm is
lower bounded by \[
1-\frac{1-\exp\left(-\alpha\right)}{\alpha-\exp\left(-\alpha\right)+\exp\left(-\alpha\left(1-p\right)\right)}\]

\end{theorem}

\begin{proof} Following the proof of the efficiency of the $C^{3}$
algorithm, we compute the rate in which the size of $G$ reduces.
However, we need to consider two settings. The first applies to teachers
that label points for the first time. The second case to consider
is teachers who label points that were previously labeled by some
other teacher. While these cases may be interleaved in time according
to algorithm $C^{4}$, w.l.o.g. we may analyze them as if they occur
in two consecutive phases.

Following Lemma~\ref{le: Q function}, teachers who label points
that were not previously labeled will leave for further process $\lab Q\left(\alpha\right)$
points out of every $\lab$ labeled points. Thus the first phase of
labeling will require $\ins$ labels and will leave $nQ\left(\alpha\right)$
points in the graph $G$. 

In the second phase, each teacher is fed with points that received
the same label by different teachers. Due to the name consistency
(\ref{eq:name consistent}) out of $l$ points that a teacher labeled
we expect $pl$ of them to have the same label due to the name consistency.
The other points are subject to contraction. From Lemma~\ref{le: Q function}
and the above argument we expect that from every $\lab$ points only
$1+\left(1-p\right)\lab Q\left(\alpha\left(1-p\right)\right)$ will
remain. The number of labels used by teachers labeling previously
labeled points is\begin{eqnarray*}
\frac{nQ\left(\alpha\right)}{1-\left(1-p\right)Q\left(\alpha\left(1-p\right)\right)-\frac{1}{\lab}}\end{eqnarray*}
 Thus, the overall number of labels used is \[
n\left(\frac{Q\left(\alpha\right)}{1-\left(1-p\right)Q\left(\alpha\left(1-p\right)\right)-\frac{1}{\lab}}+1\right)\]
 which leads to the efficiency of\begin{eqnarray*}
\lefteqn{\lim_{l\rightarrow\infty}\frac{1-\left(1-p\right)Q\left(\alpha\left(1-p\right)\right)-\frac{1}{l}}{Q\left(\alpha\right)+1-\left(1-p\right)Q\left(\alpha\left(1-p\right)\right)-\frac{1}{l}}=}\\
 &  & 1-\frac{1-\exp\left(-\alpha\right)}{\alpha-\exp\left(-\alpha\right)+\exp\left(-\alpha\left(1-p\right)\right)}\end{eqnarray*}

\end{proof}

One can easily verify, that if $p=0$ the label efficiency of the
$C^{4}$ algorithm is identical to that of the $C^{3}$ algorithm.
However, the difference between the $C^{3}$ algorithm and $C^{4}$
algorithm is profound when $p\rightarrow1$ and $\alpha\rightarrow0$.
In this setting, the $C^{3}$ algorithm has efficiency of $\left(\nicefrac{\alpha}{2}\right)+o\left(\alpha\right)$
while the $C^{4}$ algorithm is $\left(\nicefrac{1}{2}\right)-o\left(1\right)$
efficient. 

Note that despite the remarkable improvment, when $p=1$ there exists
complete name consistency and thus it is trivially possible to achieve
the perfect efficiency of $1$. However, it is not clear if it is
possible to get efficiency close to $1$ if $p$ is slightly less
than $1$. This remains as an open problem.

\section{Conclusions and further research}

In this work we have studied the problem of generating consistent
labels for a large data set given that the labels are provided by
restricted teachers. We have focused on the problems arising when
the labels used by different teachers are un-coordinated, but nevertheless
a one-to-one (unknown) correspondence exists between their labeling
systems. In this framework, we provided several algorithms and analyzed
their efficiency. We also presented an upper bound which shows that
the problem is non-trivial, and becomes hard as the number of classes
grows. In the limit $\alpha\to0$ we characterize the achievable efficiency
to be in the range%
\footnote{The representers algorithm achieves efficiency of $\left(\nicefrac{2}{3}\right)\alpha$
with $\beta=\nicefrac{1}{3}$ and $\alpha\to0$. To see this, plug
these values in (\ref{eq:representers not bounded}). %
} $\left[\left(\nicefrac{2}{3}\right)\alpha,2\alpha\right]$, however
the exact value remains as an open problem.

We believe that the process of collecting data for large scale learning
deserves much attention. One interesting extension of this work is
to the case where the symmetry between teachers is broken, either
by considering different noise levels to their labels, or more generally,
by also allowing the noise level to change between the different classes.
In such scenarios, a 'teacher selection' problem arises as the identity
of the teacher can be very informative. One example is the problem
of ``provost-selection'' in which most of the teachers are useless
novices in some domain-specific issues and thus it is essential to
first find the experts (``provosts'') and use only the labels they
provide. A related problem arises when all teachers are useful, but
they differ in their discrimination resolutions, so one teacher may
say that an image contains a bird while the other may describe the
exact bird species. Such problems are left for further research. 

{\scriptsize \bibliographystyle{mlapa}
\bibliography{rgb}
}
\end{document}